\documentclass{article}

\usepackage[final]{neurips_2026}
\makeatletter\providecommand{\@trackname}{}\makeatother

\usepackage[utf8]{inputenc}
\usepackage[T1]{fontenc}
\usepackage{hyperref}
\usepackage{url}
\usepackage{booktabs}
\usepackage{amsfonts}
\usepackage{amsmath}
\usepackage{amssymb}
\usepackage{amsthm}
\usepackage{nicefrac}
\usepackage{microtype}
\usepackage{xcolor}
\usepackage{graphicx}
\usepackage{bm}
\usepackage{fancyhdr}
\fancypagestyle{firstpage}{%
  \fancyhf{}%
  \fancyfoot[C]{\footnotesize Accepted at ICML 2026 Workshop on Decision-Making from Offline Datasets to Online Adaptation: Black-Box Optimization to Reinforcement Learning.}%
}

\theoremstyle{remark}

\newcommand{\cX}{\mathcal{X}}
\newcommand{\E}{\mathbb{E}}

\newcommand{\UCB}{\mathrm{UCB}}

\title{How Many Initial Points Does Bayesian Optimization Need?}

\author{%
  Mujin Cheon$^{1,2}$, James Odgers$^{3,4,5}$, Dong-Yeun Koh$^1$, Calvin Tsay$^6$ \\
  $^1$Korea Advanced Institute of Science \& Technology (KAIST), $^2$HayanMind Inc. \\
  $^3$ Technical University of Nuremberg (UTN), $^4$ Munich Center for Machine Learning (MCML), \\  $^5$ Helmholtz Munich,
  $^6$Imperial College London.\\
  \texttt{c.tsay@imperial.ac.uk}
}

\makeatletter
\renewcommand\paragraph{\@startsection{paragraph}{4}{\z@}%
  {0pt}%
  {-0.5em}%
  {\normalfont\normalsize\bfseries}}
\makeatother
\begin{document}

\maketitle
\thispagestyle{firstpage}

\begin{abstract}
Bayesian Optimization (BO) generally begins with an \emph{initialization phase}: a batch of $n_0$ uninformed evaluations. The choice of $n_0$ remains largely heuristic, and we empirically observe that the total cost (random initial points plus BO iterations needed to find the global optimum) is U-shaped in $n_0$, 
i.e., a practitioner wastes resources by selecting either too low or too high a value of $n_0$. We find this tradeoff persists across MLE, Bayesian MCMC, and \emph{exact} GP hyperparameters, as well as across acquisition functions. Toward the latter, Thompson Sampling appears an exception, with both total cost and simple regret essentially $n_0$-agnostic, though higher in our experiments.
We attribute this U-shape to the known boundary issue of variance-driven BO: BO burns early budget on corners of the hypercube before turning inward. We demonstrate this effect using a 3D BO trajectory where the exact hyperparameters are known. We conclude with practical recommendations: use multi-step lookahead BO where possible; otherwise use Thompson Sampling when $n_0$ cannot be tuned, and a generously large $n_0$ when it can.

\end{abstract}

\section{Introduction}
\label{sec:intro}

Sequential decision-making with expensive evaluations---whether in Bayesian Optimization (BO) for molecular design \citep{xie2025bogrape}, contextual bandits for recommendation \citep{li2010contextual}, or engineering systems design~\citep{cheon2024non, paulson2025bayesian}---routinely combines an offline phase of uninformed exploration with online, model-guided adaptation. In BO \citep{jones1998efficient, frazier2018tutorial, garnett2023bayesian}, this transition is ingrained in the standard protocol: evaluate $n_0$ initial points (by random, Latin-hypercube, or Sobol sampling), fit a Gaussian process (GP) surrogate, and then transition to using an acquisition function to guide samples. The first phase is effectively an offline dataset (and ignored in most BO literature); the second phase, where BO takes over, is online adaptation. Despite the ubiquity of this two-phase protocol, the question of how to size the offline phase has received relatively little attention. Practitioners often rely on simple rules of thumb; for instance, Meta's Ax platform \citep{olson2025ax} defaults to 5 initial trials (the search-space center plus 4 quasi-random Sobol points), independent of dimensionality. Another common rule for the GP is $n_0 = 10d$ \citep{loeppky2009choosing}, yet no analogous analysis exists for \emph{total experiment} cost.

\paragraph{The initialization tradeoff.}
Let $\tau(n_0)$ be the number of BO iterations needed to find the global optimum after $n_0$ random initial evaluations and $C(n_0) = n_0 + \E[\tau(n_0)]$ the total expected cost. We empirically observe that $C(n_0)$ is sharply U-shaped on discrete grids, and that this U-shape persists across (i) hyperparameter regimes (MLE / Bayesian MCMC / oracle) and (ii) multiple variance-driven acquisition functions.
Toward the latter, we observe that the total cost and convergence of BO with EI, UCB, and PI is sensitive to the choice of $n_0$, while Thompson Sampling~\citep{thompson1933likelihood}, or TS, is a notable exception (Section~\ref{sec:phenomenon}). We connect this phenomenon to the known \emph{boundary issue} of variance-driven BO~\citep{swersky2017improving, oh2018bock, siivola2017correcting}: GP posterior variance peaks at points farthest from observed data, and, for a box-shaped domain, those farthest points are the corners (in higher dimensions, the faces and edges). Prior work mitigates this through architectural remedies (cylindrical kernels, virtual derivatives, trust regions). In this work, we show that the simple choice of $n_0$ can shortcut the boundary itinerary.

\paragraph{Contributions.}
This work documents an initialization tradeoff in the BO total cost $C(n_0)$ on discrete grids across the various above settings.
We also show that a similar tradeoff may exist in the continuous variable setting. 
We connect this tradeoff to the known \emph{boundary issue} of variance-driven BO and formalize its mechanism, demonstrated directly on a 3D Oracle BO trajectory (Section~\ref{sec:mechanism}).
Whereas prior work addresses the boundary issue at the level of the surrogate or optimization scheme, we identify $n_0$ sizing as a simple, complementary lever. Finally, we show that a two-step lookahead acquisition can mitigate the tradeoff on the Oracle 3D setting, flattening the U-shape and pulling the optimal $n_0$ forward (Section~\ref{sec:practical}). 
We conclude with practical recommendations.

\section{The Initialization Tradeoff in Practice}
\label{sec:phenomenon}

\paragraph{Experimental setup.}
We consider three hyperparameter settings: \emph{oracle}, where the GP surrogate uses the exact generative kernel, standard \emph{MLE}, and Bayesian \emph{MCMC}. 
We investigate $C(n_0)$ across three complementary configurations (full details in Appendix~\ref{app:experiments}): \textbf{(E1)} a fixed oracle GP with Matérn-5/2 sampled on the discrete 3D grid $\{-5,\ldots,5\}^3$ using EI ($10^5$ episodes per $n_0$); \textbf{(E2)} Ackley 4D on a $9^4 {=} 6561$ grid with an RBF kernel fitted by either MLE or Bayesian NUTS-MCMC, using EI, UCB, PI, and TS ($5{,}000$ episodes; in this 4D regime, EI and LogEI
produce essentially identical optima, so we focus on EI); \textbf{(E3)} the (E1) Oracle experiment extended to four acquisition functions (EI, UCB, PI, TS), $500$ episodes per $(n_0, \text{acquisition})$ pair, $n_0 \in \{0, 10, 20, 30, 40, 50\}$.

\begin{figure}[b]
  \centering
  \includegraphics[width=\linewidth]{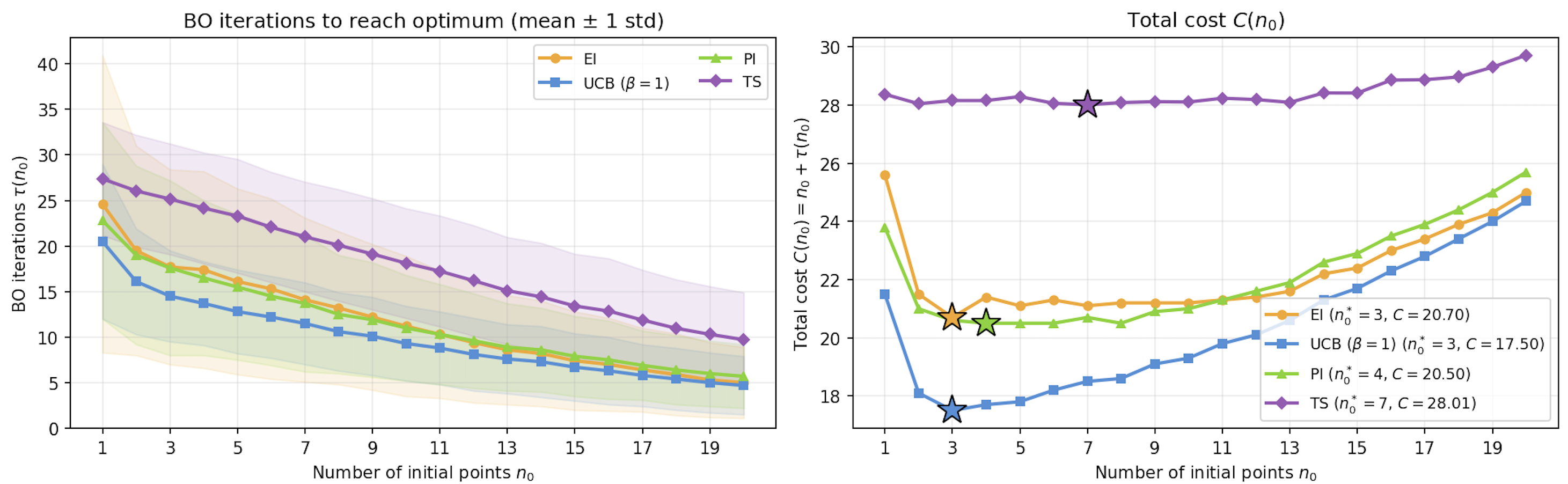}
  \caption{\textbf{(E2) MLE-fitted RBF on Ackley 4D.} Total cost $C(n_0)$ over $5{,}000$ episodes (mean $\pm$ std). This is our starting observation: the initialization tradeoff appears strongly in the most widely deployed BO setting.}
  \label{fig:ackley-mle}
\end{figure}

\begin{figure}[t]
  \centering
  \includegraphics[width=\linewidth]{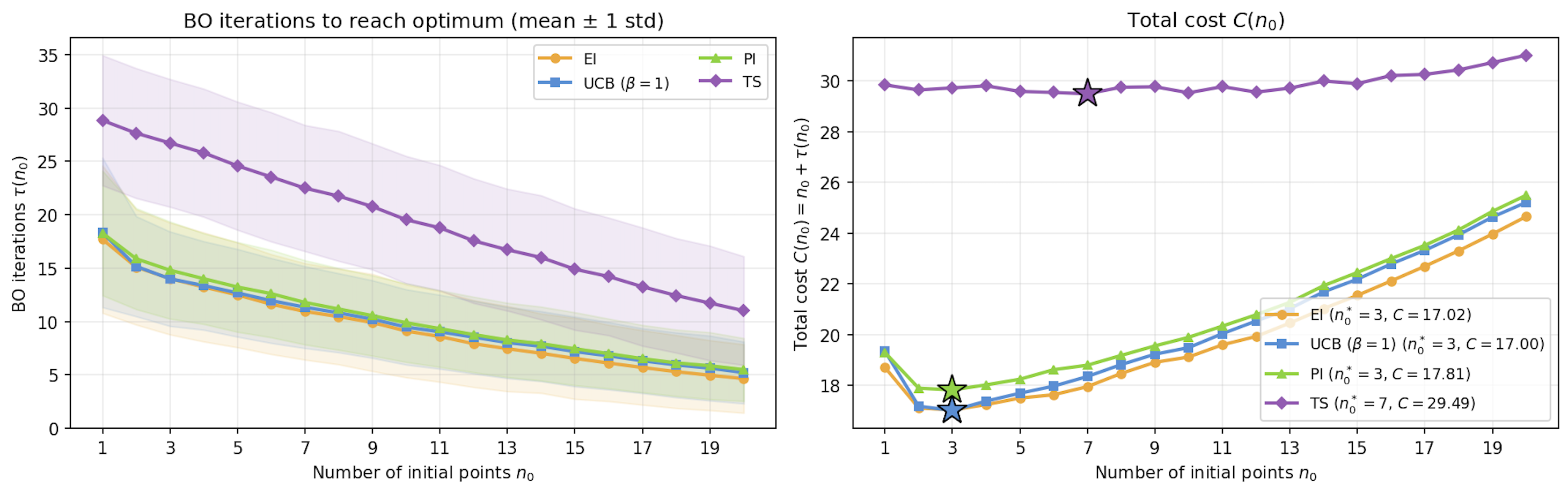}
  \caption{\textbf{(E2) Bayesian MCMC RBF on Ackley 4D.} Same setup as Figure~\ref{fig:ackley-mle} but with full kernel hyperparameter posterior. The performance gap among EI, UCB, PI collapses, but the U-shape itself stays. TS remains $n_0$-flat, at a higher cost in this setting.}
  \label{fig:ackley-mcmc}
\end{figure}

\paragraph{The initialization tradeoff is structural.}
On Ackley 4D (E2), BO with \emph{either} MLE-fitted and Bayesian MCMC-integrated kernels produces sharp U-shapes in $C(n_0)$ (Figures~\ref{fig:ackley-mle}--\ref{fig:ackley-mcmc}). 
Interestingly, while EI, UCB, and PI show acquisition-specific gaps under MLE, the three coincide closely under MCMC. In other words, accounting for hyperparameter uncertainty may mitigate differences between acquisition functions, but the U-shaped initialization tradeoff remains. 
We further test an oracle GP in (E1), leaving no mismatch between the surrogate model and underlying function, yet the tradeoff still appears (Appendix Figure~\ref{fig:e1}). The tradeoff is therefore a structural property of the \emph{decision rule}, and not a result of surrogate model error(s). The same U-shape and TS exception also persist when random initialization is replaced by Sobol or Latin-hypercube space-filling designs, which lower the overall cost but leave $n_0^\star$ and the qualitative picture unchanged (Appendix~\ref{app:init-scheme}).

\paragraph{Thompson Sampling (TS) is the lone exception.}
Figures~\ref{fig:ackley-mle}--\ref{fig:ackley-mcmc} reveal a dichotomy: EI, UCB, and PI all exhibit U-shaped $C(n_0)$ under both MLE and MCMC, whereas \emph{Thompson Sampling is the lone exception}, with $C(n_0)$ that is essentially $n_0$-agnostic, albeit at a higher absolute level in these experiments. Experiment (E3) corroborates this regret-side (Figure~\ref{fig:e3-multi-acq}; alternative plots in Appendix~\ref{app:e3-curves}), and the same pattern persists on a continuous 3D benchmark (Appendix~\ref{app:cont-acq}). Section~\ref{sec:mechanism} attributes both to a corner-by-corner boundary itinerary that random initialization shortcuts.

\begin{figure}[t]
  \centering
  \includegraphics[width=\linewidth]{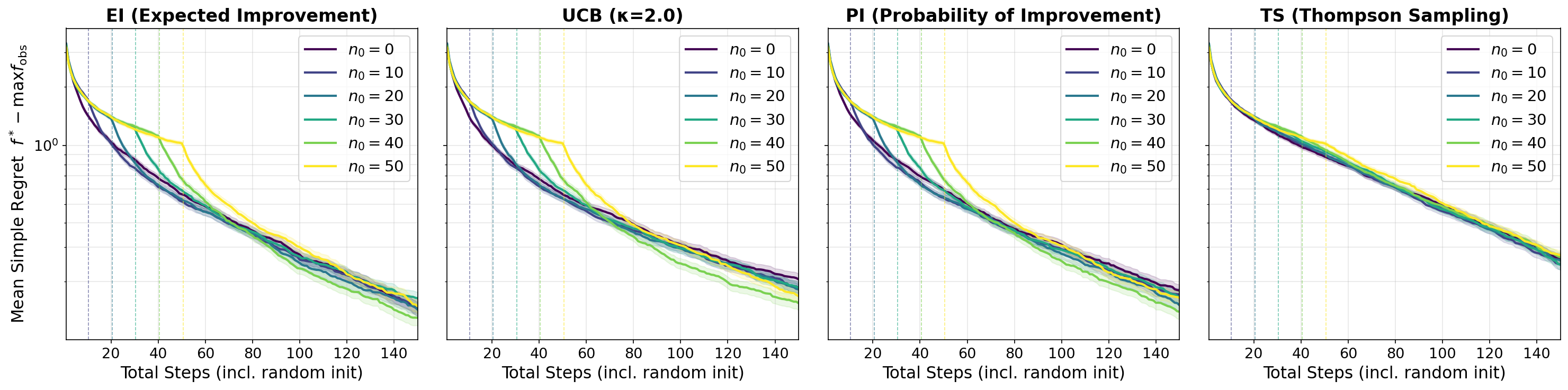}
  \caption{\textbf{(E3) Oracle 3D grid, four acquisitions.} Simple regret vs total steps on the (E1) grid $\{-5,\ldots,5\}^3$, one panel per acquisition and one curve per $n_0 \in \{0, 10, 20, 30, 40, 50\}$ (viridis; dashed = end of random init). TS (right panel) is essentially $n_0$-agnostic at a higher absolute regret.}
  \label{fig:e3-multi-acq}
\end{figure}

\section{Mechanism: Boundary Exploration Pathology}
\label{sec:mechanism}

\paragraph{Background: the boundary issue.}
The boundary issue of variance-driven BO is well documented. \citet{swersky2017improving} observes that GP posterior variance peaks at points farthest from observed data, which is likely to be on a boundary. This effect diminishes as more points are observed, which the author briefly links to initial-point density.

\paragraph{Refinement.}
Using variance-driven acquisition functions on a hypercube $[L, U]^d$ can systematically prioritize corner and face regions before the interior, a preference also underlying classical space-filling designs~\citep{johnson1990minimax}.
Consider using UCB with exploration parameter $\beta_t$:
\[
x_{n_0+1}^{\UCB} = \arg\max_{x \in \cX}
\bigl[\mu_{n_0}(x) + \beta_{n_0}^{1/2}\sigma_{n_0}(x)\bigr].
\]
In the low-data regime, the spatial variation of
$\beta_{n_0}^{1/2}\sigma_{n_0}$ likely dominates that of
$\mu_{n_0}$, and the argmax of UCB coincides with the argmax of $\sigma_{n_0}$.
EI can exhibit the same variance-dominance: in the limit
$\sigma_{n_0}(x) \to \infty$ at fixed $\mu_{n_0}(x) - f^*_{n_0}$, EI reduces to a monotone function of $\sigma_{n_0}(x)$. 

Boundary exploration is not inherently wasteful: vertices are directions of maximum posterior variance and could in principle yield important information. What makes vertex-querying pathological is the interaction between (i)~the acquisition function's preference for high-variance points and (ii)~the kernel's locality. Specifically, under stationary kernels with lengthscale $\ell \ll R$, an observation at a vertex $\bm{v}$ reduces posterior variance only within an $O(\ell)$-ball around $\bm{v}$. Therefore, sampling the boundary consumes a number of queries, scaling with the boundary complexity of $\cX$, while leaving the bulk of the search domain essentially unexplored.

Random (or pseudo-random) initialization shortcuts this itinerary: scattered initial points reduce the posterior variance everywhere (including near the corners) without the BO loop having to spend queries there. Thompson Sampling escapes the trap by construction: each step samples an independent posterior draw and selects its argmax, which is \emph{typically interior} rather than at the geometric extremes.

\begin{figure}[t]
  \centering
  \includegraphics[width=\linewidth, trim=0 9cm 0 0, clip]{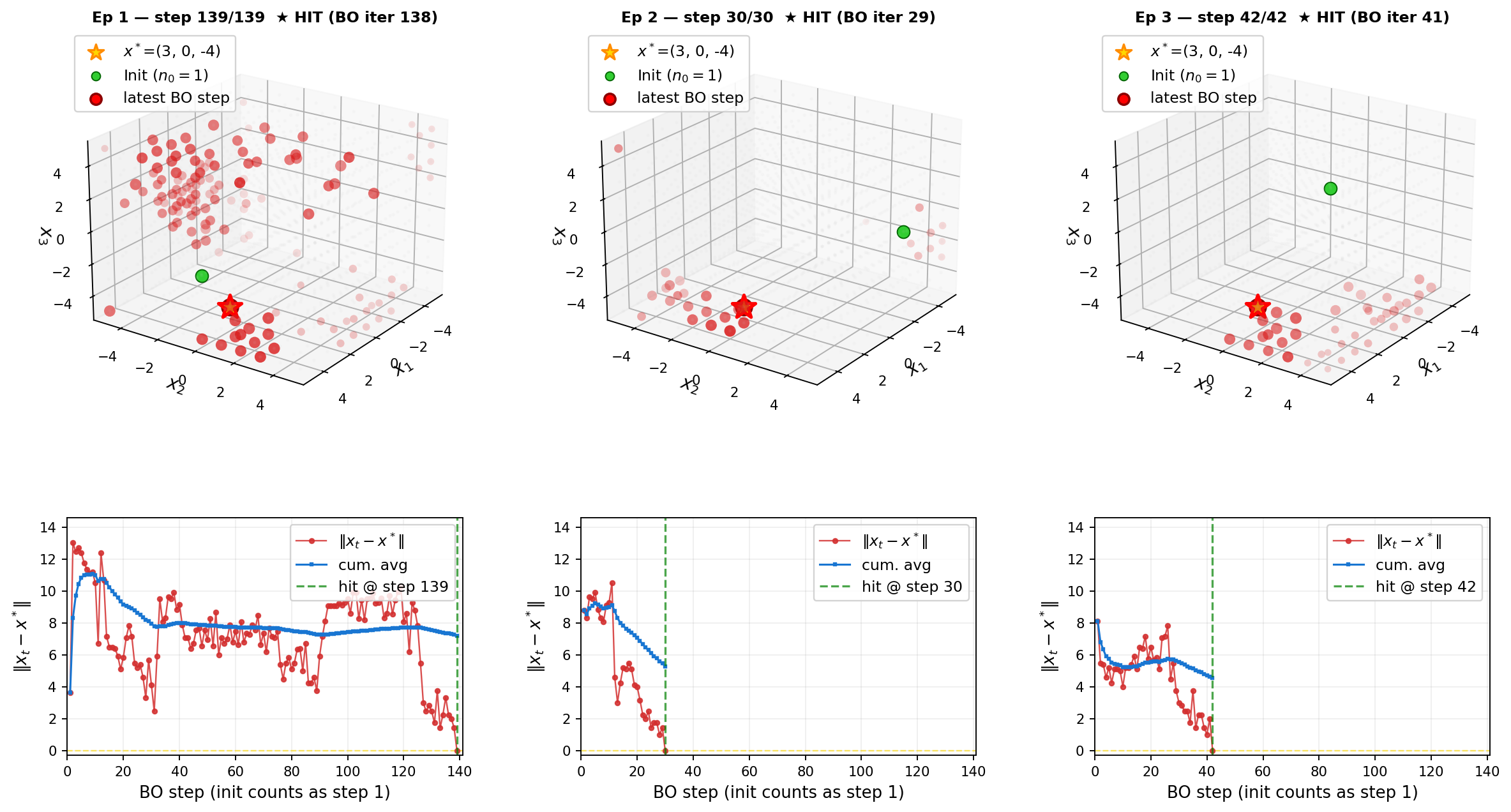}
   \includegraphics[width=\linewidth, trim=0 0 0 12cm, clip]{image/fig4_boundary_NEW.png}
  \caption{\textbf{Direct evidence for the boundary pathology (final-frame snapshot).} Three independent BO episodes on (E1) with $n_0 {=} 1$. \emph{Top row:} cumulative queries (init = green; BO trajectory = fading red, latest = solid red; $x^\star$ = gold star). \emph{Bottom row:} distance $\|x_t - x^\star\|$ vs BO step.}
  \label{fig:boundary}
\end{figure}

\paragraph{Direct evidence (3D Oracle EI, $n_0 {=} 1$).}
Figure~\ref{fig:boundary} depicts a final-frame snapshot of three independent EI-based BO procedures in setting (E1) with $n_0 {=} 1$ (same fixed $f$, $x^\star$ in the interior, three different seeds). Figure~\ref{fig:boundary} shows that BO tends to spend its early budget querying the corners farthest from the single initial observation before turning inward, even though $x^\star$ sits well in the interior. With only $n_0 {=} 1$ random observation, the surrogate is updated little from the prior, so the variance-driven boundary itinerary is exposed in its worst-case form. An unlucky initial observation pays a large boundary-exploration cost. Moderately larger $n_0$ shortcuts this itinerary, explaining the left arm of the U-shaped initialization tradeoff; eventually, however, a large $n_0$ uses too many uninformed evaluations and $C(n_0)$ rises again.

\paragraph{Discussion.}
This mechanism explains both halves of the dichotomy of Section~\ref{sec:phenomenon}: BO with variance-driven acquisitions over-explores boundaries early, giving rise to a performance tradeoff in the selection of $n_0$. On the other hand, the random sampling of TS breaks the pattern, rendering it relatively agnostic to the choice of $n_0$. The cost of this independence is that the randomized TS argmax can also miss the high-quality argmax, consistent with the higher absolute regret we observe; we report this as a property of our specific evaluations rather than a general trend, as it may partly reflect chance in these settings. We do not prove these trends formally; rather, the corner-by-corner itinerary view, together with its link to $n_0$, organizes the empirical observations of Section~\ref{sec:phenomenon} into a single causal story. Our non-oracle evidence centers on the multimodal Ackley function on a coarse grid, where the local landscape can appear smooth; a systematic comparison across unimodal geometries (e.g.\ sums of squares) and other dimensions is left to a fuller version.

\section{Practical Recommendations and Future Work}
\label{sec:practical}

\paragraph{When you cannot tune $n_0$, use Thompson Sampling.}
Our results show that TS's $C(n_0)$ (Figures~\ref{fig:ackley-mle}--\ref{fig:ackley-mcmc}) and simple-regret curves (Figure~\ref{fig:e3-multi-acq}) are both essentially independent of $n_0$. For deployments that cannot afford to tune $n_0$ ahead of time, this can be a favorable trade.

\paragraph{Otherwise, prefer a generously large $n_0$.}
The Oracle 3D U-shape is markedly asymmetric: $C(1) {\approx} 94$ and $C(35) {\approx} 81$, while $C(n_0^\star {=} 16) {\approx} 76.4$ (Figure~\ref{fig:e1}). In other words, a practitioner pays less by overshooting than by undershooting the choice of $n_0$. The boundary mechanism explains the asymmetry: too few random samples can expose the worst-case boundary itinerary, while too many merely waste a few extra uninformed samples without distorting the overall BO trajectory.

\paragraph{Multi-step lookahead BO may mitigate the tradeoff.}
The boundary pathology is myopic in nature: a single-step acquisition function does not explicitly account for the future iterations a corner query can waste. Multi-step lookahead BO strategies may implicitly price in this cost (Appendix~\ref{app:2step}).

\bibliographystyle{plainnat}
\bibliography{refs}

\clearpage
\appendix
\section{Additional Results}
\label{app:additional}

\subsection{Experimental Details}
\label{app:experiments}

This subsection expands the brief setup of Section~\ref{sec:phenomenon}.

\paragraph{(E1) Oracle GP, 3D discrete, EI.}
A single $f \sim \mathrm{GP}(0, k_{5/2})$ with lengthscale $\ell {=} 1$ and signal variance $\sigma_f^2 {=} 1$ is drawn once on $\{-5, -4, \ldots, 4, 5\}^3$ ($11^3 = 1331$ grid points). The BO surrogate uses the \emph{exact} generative kernel and hyperparameters (\texttt{optimizer\,=\,None}, \texttt{length\_scale\_bounds\,=\,fixed}); $\tau$ is the discrete first-hit time $\min\{t : x_t = x^\star\}$. We run $10^5$ episodes per $n_0 \in \{1, \ldots, 35\}$, with episode randomness coming only from the random initial subset.

\begin{figure}[h]
  \centering
  \includegraphics[width=0.55\linewidth]{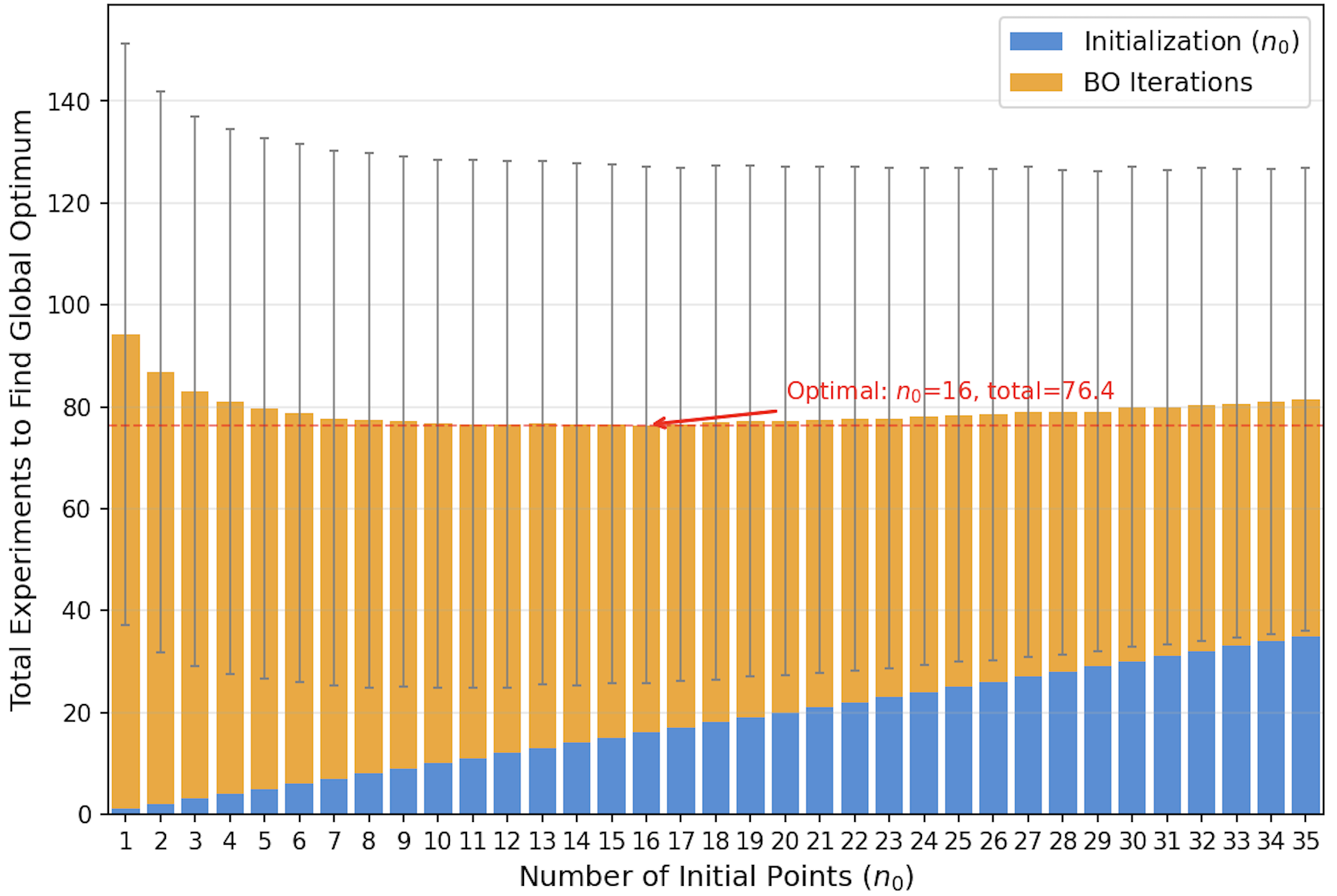}
  \caption{\textbf{(E1) Oracle 3D EI.} Total cost averaged over $10^5$ episodes with an exact-kernel oracle GP. $n_0^\star {=} 16$, $C(n_0^\star) {\approx} 76.4$; $C(1) {\approx} 94$, $C(35) {\approx} 81$.}
  \label{fig:e1}
\end{figure}

\paragraph{(E2) Ackley 4D, RBF, MLE \& MCMC, EI / UCB / PI / TS.}
$f = -\mathrm{Ackley}_4$ on $[-6, 6]^4$ discretised with grid spacing $1.5$ ($9^4 = 6561$ grid points). The surrogate uses an isotropic RBF kernel; hyperparameters are learnt either by L-BFGS MLE with $7$ restarts, or by fully Bayesian NUTS-MCMC (warmup $256$, samples $64$, target acceptance $0.9$, max tree depth $10$). $y$ values are standardised before fitting and unstandardised at prediction. $5{,}000$ episodes per $n_0 \in \{1, \ldots, 20\}$.

\paragraph{(E3) Oracle 3D grid, four acquisitions.}
Same domain, kernel, and surrogate as (E1)---fixed $f \sim \mathrm{GP}(0, k_{5/2})$ on $\{-5, \ldots, 5\}^3$ with the exact generative kernel---but extended to all four of EI, UCB ($\kappa {=} 2$), PI, and TS, with $500$ episodes per $(n_0, \text{acquisition})$ pair and $n_0 \in \{0, 10, 20, 30, 40, 50\}$. Performance is reported as simple regret $f^\star - \max f_{\mathrm{obs}}$ rather than first-hit time so that all four acquisitions can be compared on the same metric. The four acquisitions are evaluated on the \emph{same} per-episode random initializations to remove cross-acquisition variance.

\subsection{(E3) Alternative Grouping}
\label{app:e3-curves}

\begin{figure}[h]
  \centering
  \includegraphics[width=0.95\linewidth]{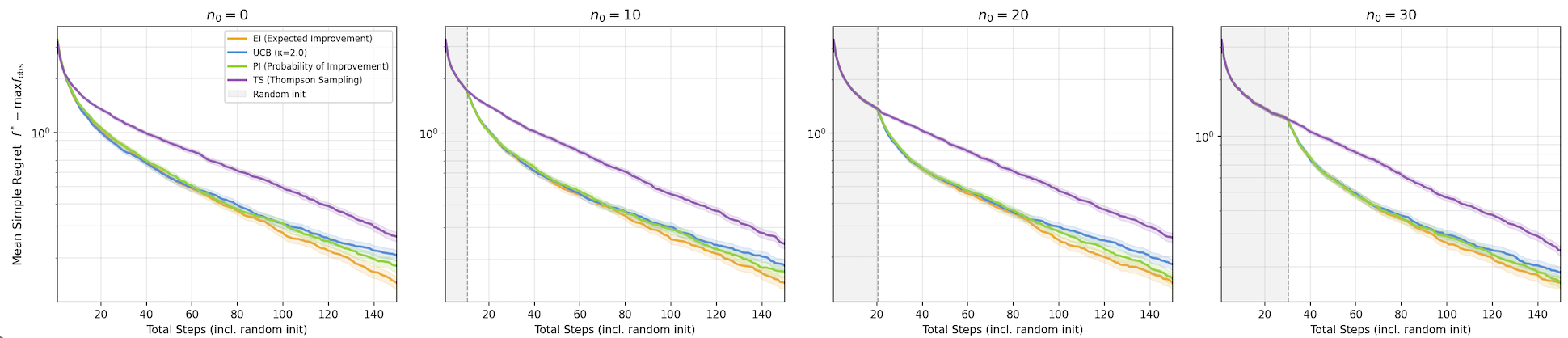}
  \caption{\textbf{(E3) Oracle 3D grid, alternative grouping.} Same data as Figure~\ref{fig:e3-multi-acq}, regrouped per $n_0$ so that the four acquisitions can be compared side by side at every $n_0$.}
  \label{fig:e3-curves}
\end{figure}

Figure~\ref{fig:e3-multi-acq} of the main text groups the (E3) data by acquisition function (one panel per acquisition, one curve per $n_0$). Figure~\ref{fig:e3-curves} below regroups the same data per $n_0$ so that the four acquisitions can be compared side by side. EI / UCB / PI cluster below TS at every $n_0$.

\subsection{Continuous 3D Corroboration}
\label{app:cont-acq}

\begin{figure}[h]
  \centering
  \includegraphics[width=0.95\linewidth]{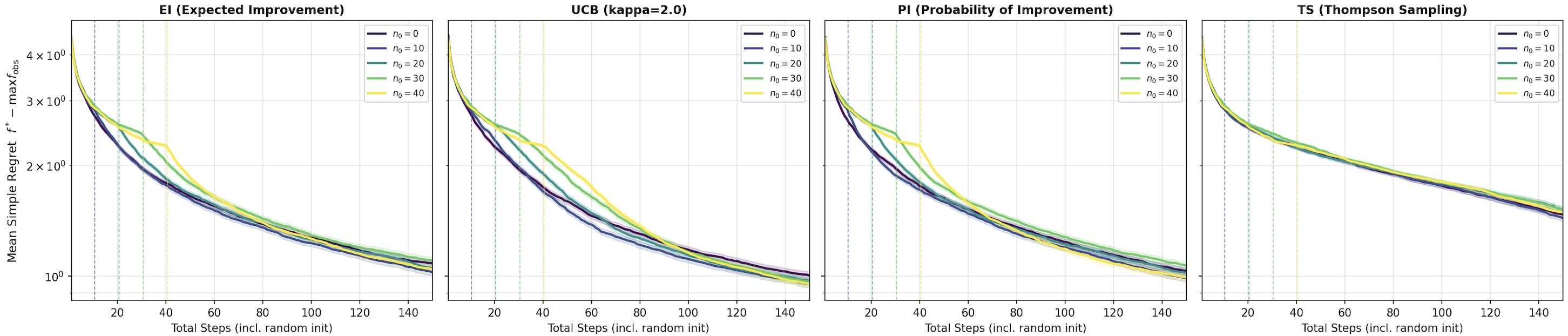}
  \caption{\textbf{Continuous 3D Oracle, four acquisitions (panel-per-acquisition view).} Simple regret vs total steps on $[-5, 5]^3$ with per-episode random Fourier feature draws, $500$ episodes per $(n_0, \text{acquisition})$ pair, $n_0 \in \{0, 10, 20, 30, 40\}$. EI / UCB / PI show a lower absolute regret with $n_0$ sensitivity; TS is essentially $n_0$-flat at a higher absolute regret.}
  \label{fig:cont-acq}
\end{figure}

\begin{figure}[h]
  \centering
  \includegraphics[width=0.95\linewidth]{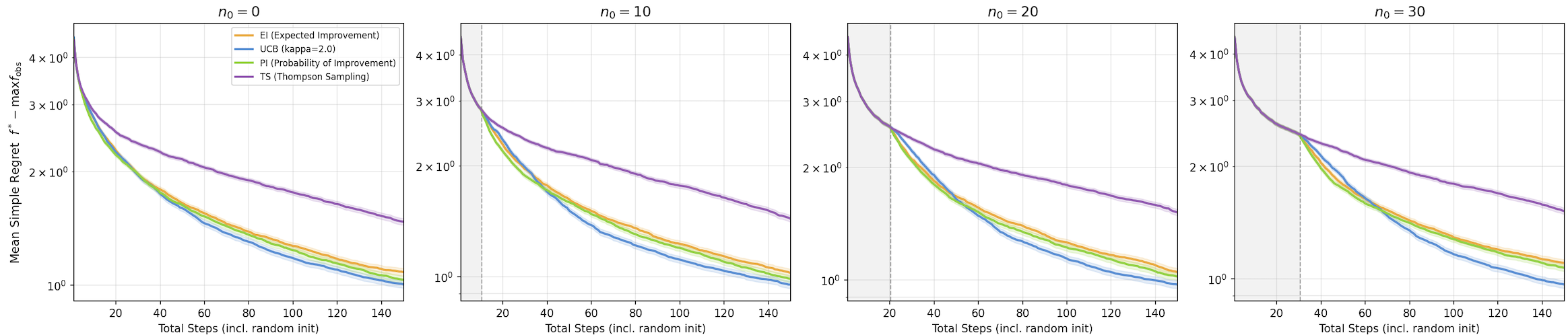}
  \caption{\textbf{Continuous 3D Oracle, alternative grouping.} Same data as Figure~\ref{fig:cont-acq}, regrouped per $n_0$ to compare the four acquisitions side by side.}
  \label{fig:cont-curves}
\end{figure}

To corroborate the (E3) regret-based dichotomy outside a fixed discrete grid, we run a continuous 3D experiment with EI, UCB ($\kappa {=} 2$), PI, and TS on per-episode random Fourier feature draws of $f$ on $[-5, 5]^3$ (candidate set re-sampled every BO step), $500$ episodes per $(n_0, \text{acquisition})$ pair, $n_0 \in \{0, 10, 20, 30, 40\}$, with matched random initializations. Figures~\ref{fig:cont-acq}--\ref{fig:cont-curves} show the same pattern as (E3): EI / UCB / PI show a lower absolute simple regret with $n_0$ sensitivity, while Thompson Sampling is essentially $n_0$-flat at a higher absolute regret level.

\subsection{Initialization Scheme: Random vs.\ Space-Filling (LHS / Sobol)}
\label{app:init-scheme}

A natural question is whether pseudo-random /space-filling initialization schemes, e.g., Sobol, Latin-hypercube (LHS), mitigate the boundary issue (and thereby change our conclusions about $n_0$). We repeat the (E2) Ackley 4D experiment with the random initial batch replaced by space-filling designs.

\paragraph{Setup.}
For each $n_0$ we draw a continuous LHS or scrambled-Sobol sample on $[-6, 6]^4$ and snap each point to its nearest grid vertex (de-duplicating, with uniform-random fill for the rare collision). All other settings match (E2) of Appendix~\ref{app:experiments}: isotropic RBF kernel, hyperparameters by L-BFGS MLE ($7$ restarts) or NUTS-MCMC, acquisitions EI / UCB / PI / TS, and $5{,}000$ episodes; here $n_0 \in \{1, \ldots, 15\}$, with MLE and MCMC sharing per-episode initial points. The random baseline is the (E2) data restricted to the same range.

\begin{table}[h]
  \centering
  \caption{\textbf{Total cost $C(n_0^\star)$ under random vs.\ space-filling initialization} (Ackley 4D, E2; TS omitted, as it is $n_0$-agnostic). Both space-filling schemes lower $C(n_0^\star)$ by $\approx 1.4$--$2$ iterations relative to random, across every acquisition and both hyperparameter regimes; LHS and Sobol are comparable (lowest in bold). Standard errors are $\lesssim 0.15$.}
  \label{tab:init-scheme}
  \begin{tabular}{lccc ccc}
    \toprule
     & \multicolumn{3}{c}{MCMC} & \multicolumn{3}{c}{MLE} \\
    \cmidrule(lr){2-4} \cmidrule(lr){5-7}
    Acq. & Random & LHS & Sobol & Random & LHS & Sobol \\
    \midrule
    EI  & 17.0 & 15.6 & \textbf{15.4} & 20.7 & 18.7 & \textbf{18.5} \\
    UCB & 17.0 & 15.7 & \textbf{15.6} & 17.5 & 16.0 & \textbf{15.8} \\
    PI  & 17.8 & 16.5 & \textbf{16.4} & 20.5 & 18.6 & \textbf{18.3} \\
    \bottomrule
  \end{tabular}
\end{table}

\paragraph{Space-filling lowers the level, not the location.}
Two patterns hold across EI, UCB, PI and both hyperparameter regimes (Table~\ref{tab:init-scheme}). First, both space-filling schemes lower the total cost relative to random initialization by roughly $1.4$--$2$ iterations, with LHS and Sobol performing comparably (Sobol is marginally lower). This is what the boundary mechanism of Section~\ref{sec:mechanism} predicts: a space-filling batch reduces posterior variance near the boundary up front more evenly than random scatter. We note this behavior mirrors the rationale of classical space-filling designs~\citep{johnson1990minimax}. Second, the \emph{location} of the optimum is essentially unchanged, $n_0^\star \approx 3$ under all three schemes. (The MLE curves are noisier, so their argmin is less stable; but $n_0 {=} 3$ is statistically indistinguishable from the minimum in every MLE cell.) In short, space-filling shifts the U-shape \emph{down} rather than \emph{left}.

\begin{figure}[b]
  \centering
  \includegraphics[width=\linewidth]{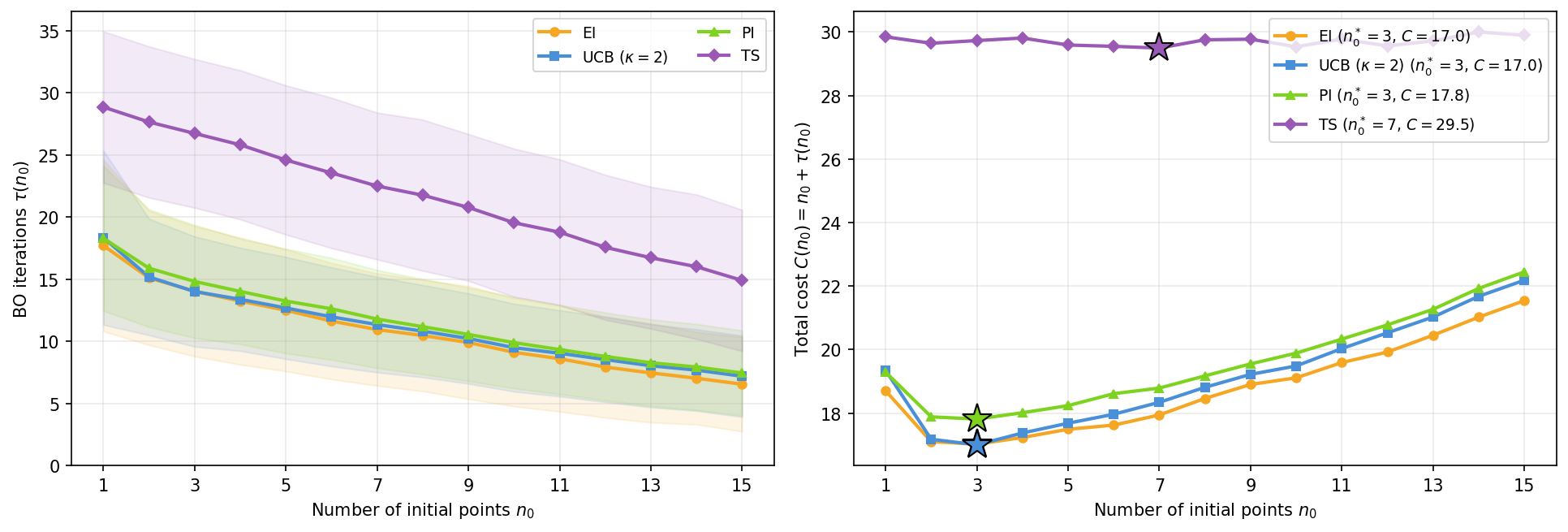}\\[2pt]
  \includegraphics[width=\linewidth]{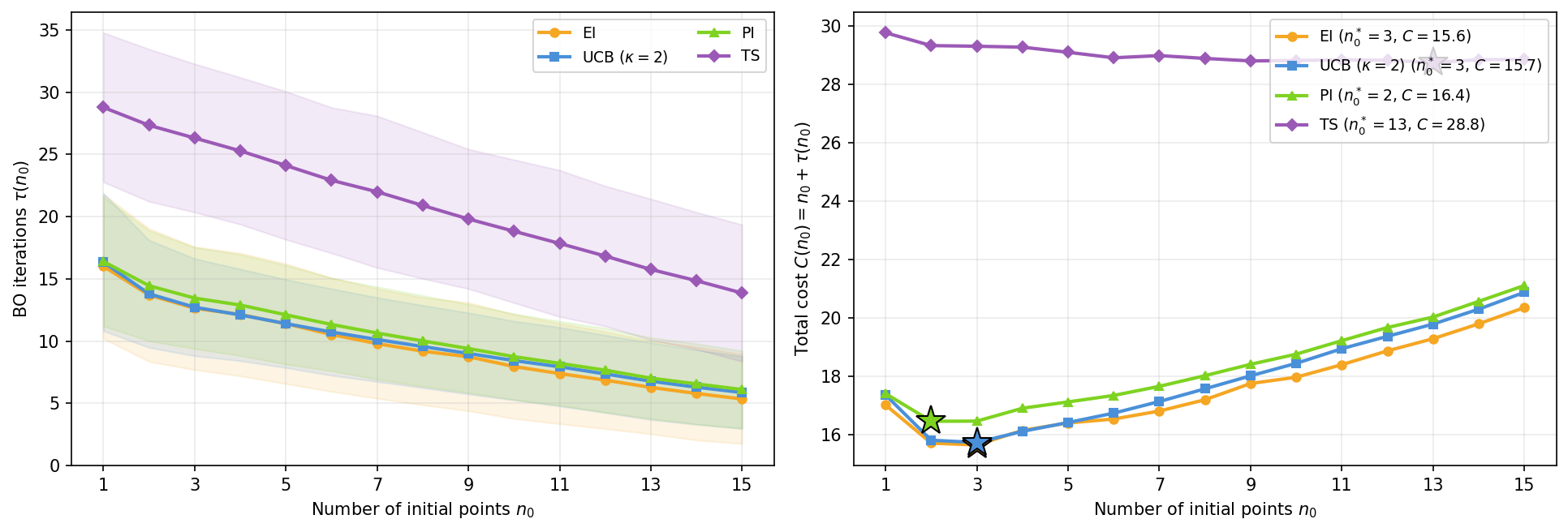}\\[2pt]
  \includegraphics[width=\linewidth]{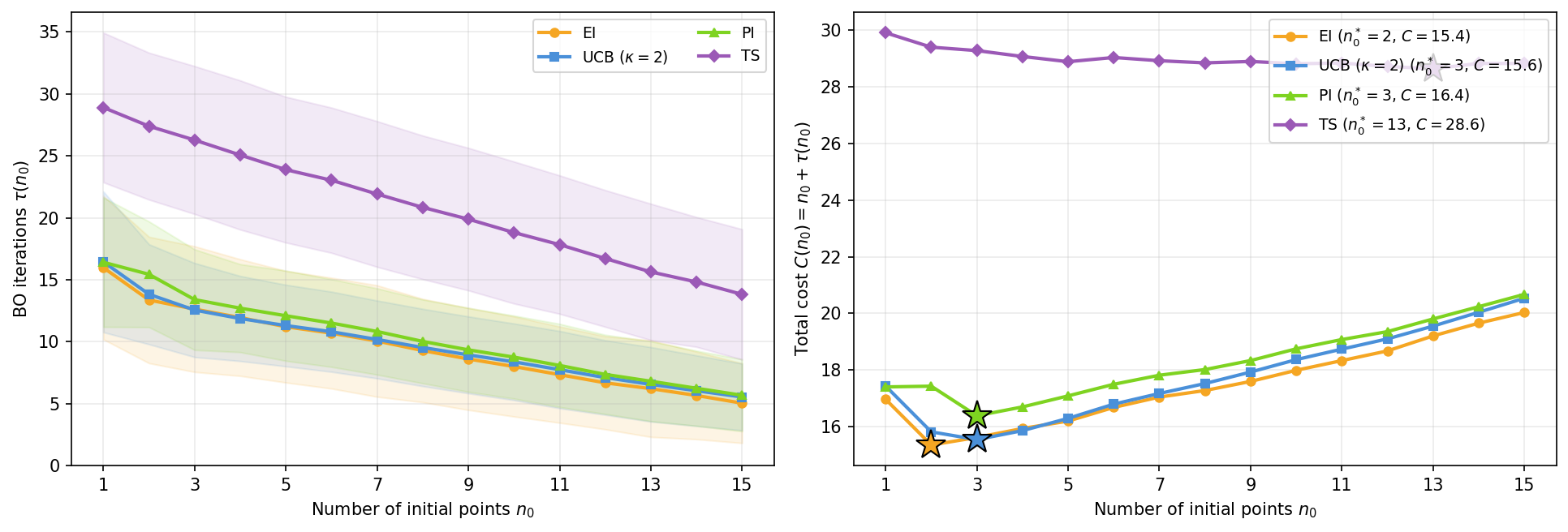}
  \caption{\textbf{(E2) Ackley 4D, MCMC: four acquisitions under each initialization scheme.} Top to bottom: random, LHS, Sobol. \emph{Left:} BO iterations $\tau(n_0)$ to reach the optimum (mean $\pm$ 1 std). \emph{Right:} total cost $C(n_0) = n_0 + \tau(n_0)$, with $n_0^\star$ starred. EI/UCB/PI stay U-shaped and TS stays $n_0$-flat under all three schemes; the EI/UCB/PI cost level drops from random to LHS/Sobol while $n_0^\star \approx 3$ is unchanged. The random panels are the (E2) baseline (cf.\ Figures~\ref{fig:ackley-mle}--\ref{fig:ackley-mcmc}), here on $n_0 \in [1, 15]$.}
  \label{fig:acq-by-init-mcmc}
\end{figure}

\begin{figure}[b]
  \centering
  \includegraphics[width=\linewidth]{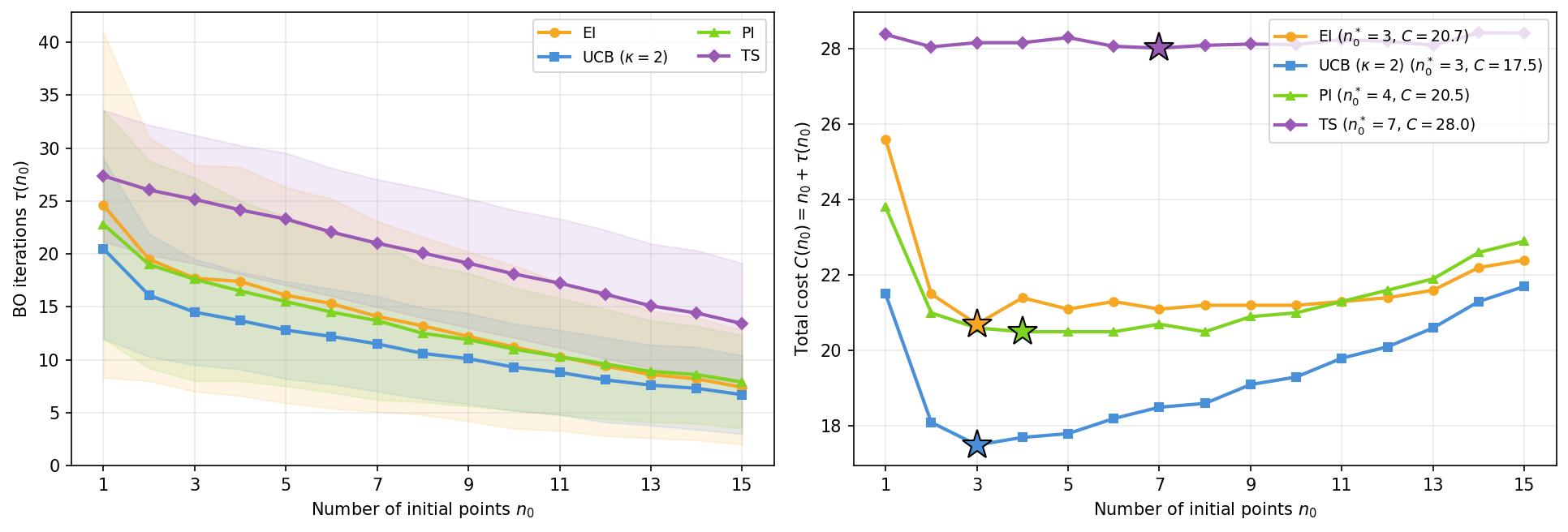}\\[2pt]
  \includegraphics[width=\linewidth]{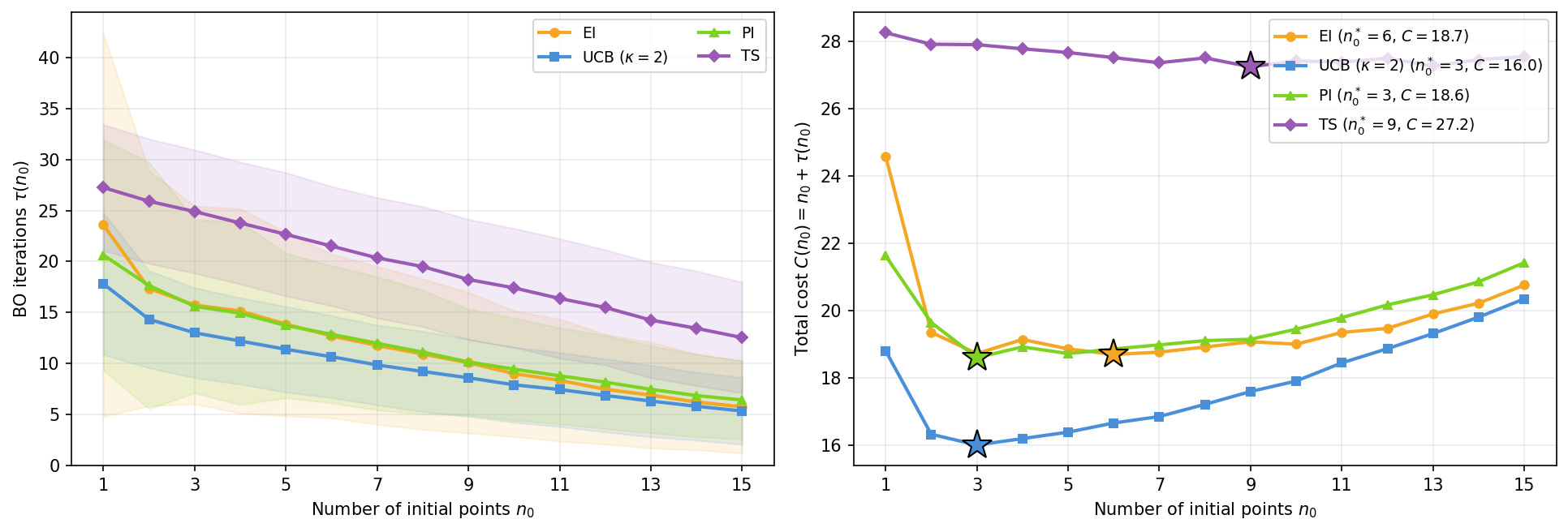}\\[2pt]
  \includegraphics[width=\linewidth]{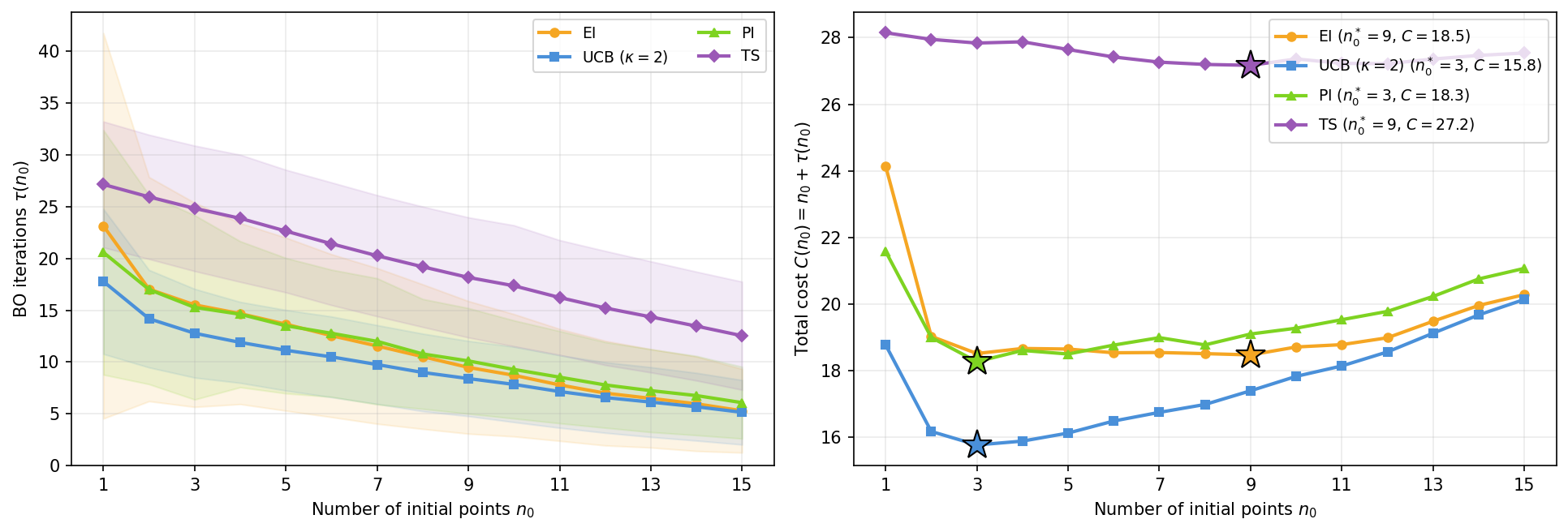}
  \caption{\textbf{(E2) Ackley 4D, MLE: four acquisitions under each initialization scheme.} Top to bottom: random, LHS, Sobol. Axes as in Figure~\ref{fig:acq-by-init-mcmc}. The MLE curves are noisier than their MCMC counterparts, so the argmin location is less stable, but $n_0 {=} 3$ remains statistically indistinguishable from the minimum throughout; the qualitative picture (EI/UCB/PI U-shaped, TS flat, level lowered by space-filling) is unchanged.}
  \label{fig:acq-by-init-mle}
\end{figure}

\paragraph{The core conclusions are unchanged.}
The two headline findings of Section~\ref{sec:phenomenon} are consistent to the choice of initialization strategy (results in Figures~\ref{fig:acq-by-init-mcmc}--\ref{fig:acq-by-init-mle}): EI, UCB, and PI remain U-shaped in $n_0$, and TS remains essentially $n_0$-agnostic. Replacing random with Sobol or LHS initialization reduces the boundary-exploration cost, but does not alter our recommendations on $n_0$ sizing or the TS exception.

\paragraph{Conjecture: room to move in harder problems.}
On this 4D grid the optimum already sits at a small $n_0^\star \approx 3$, so there is little room for a space-filling design to move it further left, and we accordingly observe a change in level rather than location. The optimal value is, however, problem-dependent and can be far larger (e.g., $n_0^\star \approx 16$ in the higher-cost (E1) oracle setting, Figure~\ref{fig:e1}). We therefore conjecture that in harder, higher-dimensional problems---where $n_0^\star$ is substantially larger---space-filling initialization may additionally shift $n_0^\star$ \emph{leftward}, reducing the number of initial points required, rather than merely lowering $C(n_0^\star)$. Testing this requires the higher-dimensional, varied-geometry benchmarks we leave to future work.

\subsection{Two-Step Lookahead on the Oracle 3D Grid}
\label{app:2step}

Multi-step lookahead BO strategies may implicitly price in the cost of wasteful initial queries. Three complementary methods are immediate candidates: \citet{cheonearl}'s EARL-BO uses reinforcement learning for high-dimensional multi-step BO, multi-step expected improvement (MSEI) admits efficient optimization via one-shot multi-step trees \citep{jiang2020efficient}, and \citet{wu2019practical} provide a practical two-step lookahead acquisition. As preliminary evidence that this direction can help, this section shows a two-step lookahead variant flattens the initialization tradeoff on our Oracle 3D setting.
Specifically, we repeat the (E1) Oracle 3D experiment with a
two-step lookahead EI acquisition in place of single-step EI---the same
fixed-kernel oracle GP on $\{-5, \ldots, 5\}^3$, here over $50{,}000$ episodes
per $n_0$. 

\begin{figure}[b]
  \centering
  \includegraphics[width=0.95\linewidth]{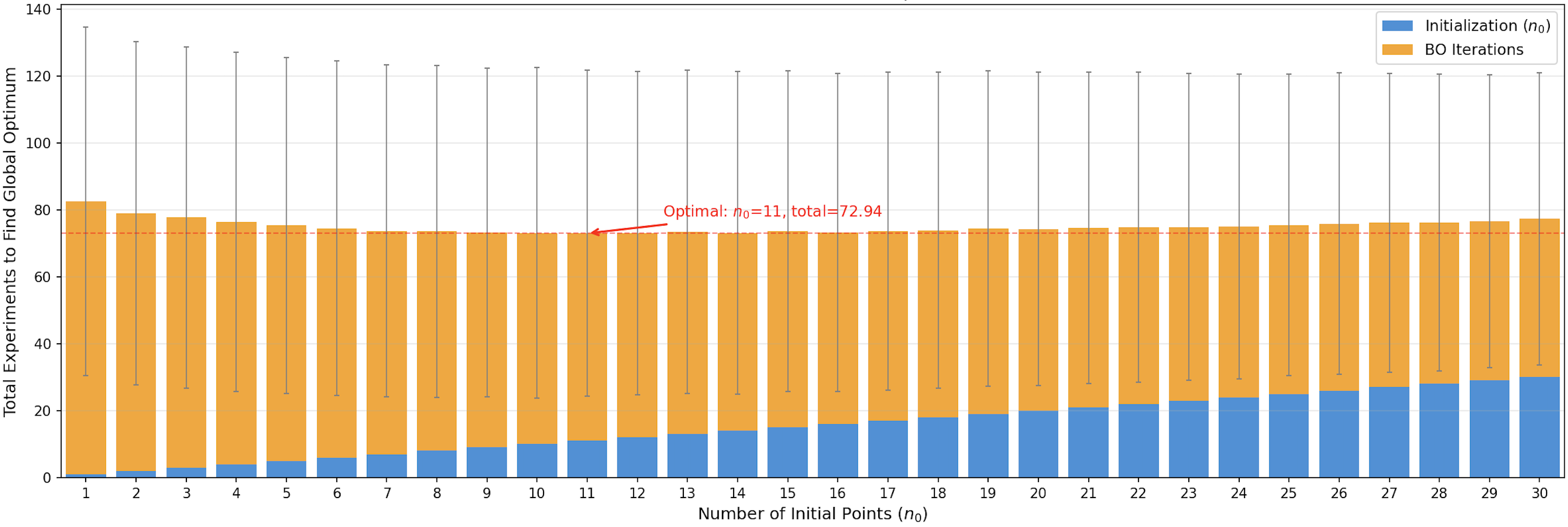}
  \caption{\textbf{Two-step lookahead EI on the (E1) Oracle 3D grid.} Total experiments to reach the optimum (initialization $n_0$, blue; BO iterations, orange) versus $n_0$, over $50{,}000$ episodes. The optimum is at $n_0^\star {=} 11$ ($C {=} 72.94$), and the U-shape is markedly flatter than the single-step EI baseline of Figure~\ref{fig:e1} ($n_0^\star {\approx} 16$, $C {\approx} 76.4$).}
  \label{fig:2step}
\end{figure}

The results of this experiment are shown in Figure~\ref{fig:2step}. 
Compared with single-step EI ($n_0^\star {\approx} 16$,
$C {\approx} 76.4$; Figure~\ref{fig:e1}), two-step EI flattens the U-shape and
moves the optimum forward to $n_0^\star {=} 11$ with $C {\approx} 72.9$. This is consistent with our myopia hypothesis above: looking even one step ahead prices in the future cost of a boundary query, so performance depends much less on a well-sized initial design. We interpret this as preliminary evidence that two-step lookahead can help
mitigate the initialization tradeoff, while preserving the advantages of
variance-driven acquisitions. A broader evaluation across functions and dimensions is left to future
work. Where the added acquisition-optimization cost of non-myopic strategies is acceptable, they may provide practical benefits for the initialization phase.

\end{document}